\pdfoutput=1

\documentclass[11pt]{article}

\usepackage[preprint]{acl}
\usepackage{multirow}
\usepackage{makecell}
\usepackage{graphicx}
\usepackage{booktabs}  
\usepackage{amsmath, amssymb}
\usepackage{xcolor}
\usepackage{booktabs}   
\usepackage{colortbl}   
\usepackage[textsize=small]{todonotes}
\usepackage{times}
\usepackage{latexsym}

\usepackage[T1]{fontenc}

\usepackage[utf8]{inputenc}

\usepackage{microtype}

\usepackage{inconsolata}

\usepackage{graphicx}
\usepackage{listings}
\definecolor{codebg}{rgb}{0.95,0.95,0.95}
\definecolor{modernbg}{rgb}{0.95,0.95,1}
\definecolor{modernborder}{rgb}{0.6,0.6,1}

\newcommand{\intrametric}{\texttt{ITF-IDF}}
\newcommand{\topic}{\texttt{LLMs Assist NLP Researchers}}
\newcommand{\dataname}{\texttt{ReviewCritique}}

\newcommand{\llmasreviewers}{\texttt{LLMs as Reviewers}}
\newcommand{\llmasmeta}{\texttt{LLMs as Metareviewers}}
\newcommand{\unreliable}{\texttt{Deficient}}

\newcommand{\newmetric}{\texttt{ITF-IDF}}
\newcommand{\promptall}{\texttt{Labeling-All}}
\newcommand{\promptneg}{\texttt{Select-\unreliable}}

\title{\topic: Critique Paper (Meta-)Reviewing}

\author{
 \textbf{Jiangshu Du\textsuperscript{1}},
 \textbf{Yibo Wang\textsuperscript{1}},
 \textbf{Wenting Zhao\textsuperscript{1,10}},
 \textbf{Zhongfen Deng\textsuperscript{1}},
 \textbf{Shuaiqi Liu\textsuperscript{3}},
 \\
 \textbf{Renze Lou\textsuperscript{2}},
 \textbf{Henry Peng Zou\textsuperscript{1}},
 \textbf{Pranav Narayanan Venkit\textsuperscript{2}},
 \textbf{Nan Zhang\textsuperscript{2}},
 \textbf{Mukund Srinath\textsuperscript{2}},
 \\
 \textbf{Ranran Haoran Zhang\textsuperscript{2}},
 \textbf{Vipul Gupta\textsuperscript{2}},
 \textbf{Yinghui Li\textsuperscript{4}},
 \textbf{Tao Li\textsuperscript{5}},
 \textbf{Fei Wang\textsuperscript{6}},
 \textbf{Qin Liu\textsuperscript{7}},
 \textbf{Tianlin Liu\textsuperscript{8}},
 \\
 \textbf{Pengzhi Gao\textsuperscript{9}},
 \textbf{Congying Xia\textsuperscript{10}},
 \textbf{Chen Xing\textsuperscript{11}},
 \textbf{Cheng Jiayang\textsuperscript{12}},
 \textbf{Zhaowei Wang\textsuperscript{12}},
 \textbf{Ying Su\textsuperscript{12}},
 \\
 \textbf{Raj Sanjay Shah\textsuperscript{13}},
 \textbf{Ruohao Guo\textsuperscript{13}},
 \textbf{Jing Gu\textsuperscript{14}},
 \textbf{Haoran Li\textsuperscript{15}},
 \textbf{Kangda Wei\textsuperscript{16}},
 \textbf{Zihao Wang\textsuperscript{12}},
 \\
 \textbf{Lu Cheng\textsuperscript{1}},
 \textbf{Surangika Ranathunga\textsuperscript{17}},
 \textbf{Meng Fang\textsuperscript{18}},
 \textbf{Jie Fu\textsuperscript{12}},
 \textbf{Fei Liu\textsuperscript{21}},
 \textbf{Ruihong Huang\textsuperscript{16}},
 \\
 \textbf{Eduardo Blanco\textsuperscript{19}},
 \textbf{Yixin Cao\textsuperscript{20}},
 \textbf{Rui Zhang\textsuperscript{2}},
 \textbf{Philip S. Yu\textsuperscript{1}},
 \textbf{Wenpeng Yin\textsuperscript{2}}
\\
\\
 \textsuperscript{1}University of Illinois Chicago,
 \textsuperscript{2}Penn State University,
 \textsuperscript{3}Hong Kong Polytechnic University,
 \\
 \textsuperscript{4}Tsinghua Univeristy,
 \textsuperscript{5}Google DeepMind,
 \textsuperscript{6}University of Southern California,
 \\
 \textsuperscript{7}University of California, Davis,
 \textsuperscript{8}University of Basel, Switzerland,
 \textsuperscript{9}Xiaomi AI Lab,
 \\
 \textsuperscript{10}Salesforce Research,
 \textsuperscript{11}Scale AI,
 \textsuperscript{12}HKUST,
 \textsuperscript{13}Georgia Institute of Technology,
 \\
 \textsuperscript{14}University of California, Santa Cruz,
 \textsuperscript{15}Singapore University of Technology and Design,
 \\
 \textsuperscript{16}Texas A\&M University,
 \textsuperscript{17}Massey University, New Zealand,
 \textsuperscript{18}University of Liverpool,
 \\
 \textsuperscript{19}University of Arizona,
 \textsuperscript{20}Fudan University,
 \textsuperscript{21}Emory University
\\
\texttt{jdu25@uic.edu}, \texttt{wenpeng@psu.edu}
}

\begin{document}
\maketitle
\begin{abstract}

\textcolor{red}{\textbf{Claim}: This work is not advocating the use of LLMs for paper (meta-)reviewing. Instead, we present a comparative analysis to identify and distinguish LLM activities from human activities. Two research goals: i) Enable better recognition of instances when someone implicitly uses LLMs for reviewing activities; ii) Increase community awareness that LLMs, and AI in general, are currently inadequate for performing tasks that require a high level of expertise and nuanced judgment. }

This work is motivated by two key trends. On one hand, large language models (LLMs) have shown remarkable versatility in various generative tasks such as writing, drawing, and question answering, significantly reducing the time required for many routine tasks. On the other hand, researchers, whose work is not only time-consuming but also highly expertise-demanding, face increasing challenges as they have to spend more time reading, writing, and reviewing papers. This raises the question: how can LLMs potentially assist researchers in alleviating their heavy workload?

This study focuses on the topic of \topic, particularly examining the \textit{effectiveness} of LLM in assisting paper (meta-)reviewing and its \textit{recognizability}. To address this, we constructed the \dataname~dataset, which includes two types of information: (i) NLP papers (initial submissions rather than camera-ready) with both human-written and LLM-generated reviews, and (ii) each review comes with ``deficiency'' labels and corresponding explanations for individual segments, annotated by experts. Using \dataname, this study explores two threads of research questions: (i) ``\llmasreviewers'', how do reviews generated by LLMs compare with those written by humans in terms of quality and distinguishability? (ii) ``\llmasmeta'', how effectively can LLMs identify potential issues, such as \unreliable~or unprofessional review segments, within individual paper reviews? To our knowledge, this is the first work to provide such a comprehensive analysis. Our dataset is available at \href{https://github.com/jiangshdd/ReviewCritique}{https://github.com/jiangshdd/ReviewCritique}.

\end{abstract}

\section{Introduction}
Artificial intelligence (AI), particularly through the recent development of large language models (LLMs), has demonstrated remarkable versatility in tasks such as writing, drawing, and question answering~\cite{naveed2023comprehensive, rasool2024evaluating, kaddour2023challenges}. This has led to significant automation of many time-consuming jobs, potentially replacing more roles with AI. {Interestingly, while researchers, the creators of AI/LLMs, benefit from LLMs for simple tasks~\cite{meyer2023chatgpt, altmae2023artificial}, it still takes years to train a qualified researcher due to the domain-specific and expertise-demanding nature of their work.} Researchers now face increasing challenges with more papers to read, to beat, to write, and to review, resulting in longer and more intensive work hours. This raises the question: how promising is the potential for LLMs to work as researchers to alleviate their heavy and somewhat unhealthy workload? 

Within the scope of \topic, this work focuses on how well LLMs can perform (meta-)reviewing. 
{AI-related conferences and journals are seeing a rapid increase in submissions, making it difficult to recruit enough (meta-)reviewers. Paper reviewers must carefully read submissions and provide comments on the overall story, strengths,  weaknesses, writing, etc. The meta-reviewer's responsibility is to ensure the accuracy and constructiveness of the individual review. Therefore, meta-reviewers are expected to be aware of the submission as well as authors' rebuttals, and then assess individual reviews by identifying unreasonable elements and distilling truly constructive comments. There is a latent trend, though debatable and unacknowledged by reviewers, of LLMs participating more frequently in the paper-reviewing process.} Therefore, this work explores two research questions: (i) `\llmasreviewers'', how far away or distinguishable are LLM-generated paper reviews from human-written ones? (ii) ``\llmasmeta'', can LLMs identify \unreliable~review segments by reasoning over the paper submission, other individual reviews, and author rebuttals jointly?

To achieve this, we create the \dataname~dataset, containing: (i) NLP papers (original submissions rather than the final camera-ready) with both human-written and LLM-generated reviews, and (ii) each review annotated by NLP experts (most with Ph.D. degrees or area chairing experience) at the sentence level regarding deficiency and professionalism, with explanations. This dataset enables the following analyses.

First, for \llmasreviewers, we assess the quality of LLM-generated reviews by examining subsections or aspects of the review, such as summary, strengths, weaknesses, writing, etc. We propose a novel metric to measure LLM-generated review diversity across different papers. Our findings indicate that LLMs generate more \unreliable~review segments than human reviewers and often produce paper-unspecific reviews lacking diversity and constructive feedback.

Second, for \llmasmeta, we evaluate LLMs' ability to identify \unreliable~segments in human-written reviews and provide explanations for their judgments. This contrasts with other works treating paper meta-review as a text summarization task given 3+ individual reviews~\cite{li2023summarizing, shen-etal-2022-mred, PRADHAN2021218}. We argue that meta-reviewing should be a knowledge-intensive and reasoning-intensive process, with human meta-reviewers being expected to be careful and responsible. We benchmark both closed-source and open-source LLMs on this task, finding that even top-tier LLMs struggle to mimic human experts in assessing individual reviews.

Overall, our contributions are threefold: (i) the \dataname~dataset with human-written and LLM-generated reviews and fine-grained review deficiency labeling and explanation, serving as a valuable resource for future research on AI-assisted peer review and LLM benchmarking, (ii) the first quantitative comparison of human-written and LLM-generated paper reviews at the sentence level, and (iii) the first analysis of LLMs' potential as both reviewers and meta-reviewers. By highlighting the strengths and limitations of LLMs in scientific peer review, our work paves the way for future works on integrating AI for research.

\section{Related Work}
Researchers have explored various aspects of AI for reviews.
One area of interest is the use of AI to assist in automatically generating peer reviews, such as predicting scores \cite{li-etal-2020-multi-task, zhou2024llm, wang-etal-2020-reviewrobot, deng2020hierarchical} and writing reviews \cite{gao2024reviewer2, wang-etal-2020-reviewrobot, yuan2022can_asap, liu2023reviewergpt} and meta-reviews \cite{li2023summarizing, lin2023moprd}. Another line of research focuses on leveraging NLP methods to evaluate the quality of human reviews \cite{xiong-litman-2011-automatically, guo2023automatic, kumar2023reviewers, ghosal2022hedgepeer}.

To facilitate research on AI for peer review, several datasets have been introduced. PeerRead \cite{kang-etal-2018-dataset}, MOPRD \cite{lin2023moprd}, and NLPeer \cite{dycke-etal-2023-nlpeer} are datasets containing a large number of peer reviews and their corresponding papers but without expert annotations. 
Other datasets focus on specific aspects of peer reviews, such as argument~\cite{kennard-etal-2022-disapere, hua-etal-2019-argument, yuan2022can_asap, cheng2020ape, ruggeri2023dataset}, politeness \cite{bharti2023politepeer}, uncertainty detection \cite{ghosal2022hedgepeer}, contradictions in review pairs \cite{kumar2023reviewers}, and substantiation \cite{guo2023automatic}. Peer Review Analyze \cite{ghosal2022peer} annotates reviews across four facets: paper section correspondence, aspect, functionality, and significance. However, these datasets are solely based on reviews and none of them are \textit{highly} expert-demanding. In contrast, \dataname~ is the first dataset to benchmark LLMs' capability as a responsible meta-reviewer.

Recently, researchers have also explored the evaluation of LLMs' deficiency and limitations in automatic paper reviewing tasks \cite{zhou2024llm, liu2023reviewergpt, robertson2023gpt4, liang2023can, LIN2023101830}. Our work differs from previous works in that we provide a quantitative comparison of human-written and LLM-generated paper reviews at the sentence level. This fine-grained analysis allows us to identify specific areas where LLMs excel or struggle in generating high-quality reviews. We also propose a novel metric to measure LLM-generated review diversity.
\section{\dataname~Curation}
\label{sec:data_collection}

In this section, we detail the process of curating \dataname, including the criteria for paper selection, the collection of human-written and LLM-generated reviews, the annotation procedure, and the measures taken to ensure data quality.

\subsection{Paper Submission \& Review Collection}

\paragraph{Criteria.} We select the papers based on the following criteria: i) Only consider NLP papers; this facilitates the recruitment of sufficient annotators in the NLP domain.
ii) Human-written reviews are publicly accessible.
iii) Equal distribution of accepted and rejected papers is maintained to investigate potential review pattern discrepancies based on the final acceptance or rejection of submissions.

From the OpenReview website, we gathered 100 NLP papers (submitted to top-tier AI conferences ICLR and NeurIPS between 2020 and 2023) along with their complete individual reviews (3-5 for each submission), meta-reviews, and author rebuttals. The revision history on OpenReview allowed us to collect the latest paper submissions before the conference deadline, as these versions are the ones on which the reviews are based.
 
\textbf{Question: How can we ensure that the collected individual reviews are written by human experts rather than AI?} During the subsequent annotation process, we instruct annotators to notify us if they suspect that a review collected here was likely generated by AI; if any doubts arise, we will discard the paper and all its metadata.

\paragraph{Collecting LLM-generated Reviews} To directly compare human-written and LLM-generated reviews, we selected a subset of 20 papers from the original 100. The main reason for this selection was the time-consuming nature of subsequent annotation; a size of 20 allowed for an acceptable statistical comparison. This subset of papers also maintains an equal distribution of accepted and rejected papers. We utilized three of the most powerful closed-source LLMs, namely GPT-4~\cite{openai2023gpt4}, Gemini-1.5~\cite{google2023gemini}, and Claude Opus~\cite{claude}, as these are the models most likely to be used by humans seeking AI assistance in their reviews. Each LLM generated three reviews using prompts that included the ICLR review guidelines, randomly chosen human-written reviews for both accepted and rejected papers, and a generation template in ICLR 2024 format. This prompt can be found in Table~\ref{tab:llm_generate_revivew_prompt} (Appendix~\ref{appendix_prompt_templates}).

\begin{table*}[!ht]

    \centering
    \resizebox{0.85\linewidth}{!}{
    \begin{tabular}{lrrrrrr}
        \toprule
         & \multicolumn{3}{c}{\textbf{Human-written Review}} & \multicolumn{3}{c}{\textbf{LLM-generated Review}} \\
        \cmidrule(lr){2-4} \cmidrule(lr){5-7}
        & \textbf{All} & \textbf{Accepted} & \textbf{Rejected} & \textbf{All} & \textbf{Accepted} & \textbf{Rejected} \\
        \midrule
        \#Papers & 100 & 50 & 50 & 20 & 10 & 10 \\
        \#Reviews & 380 & 195 & 185 & 60 & 30 & 30 \\
        \enspace\enspace w/ \unreliable~seg. & 272 & 132 & 140 & 60 & 30 & 30 \\
        \enspace\enspace w/ \unreliable~pct. (\%) & 71.57 & 67.69 & 75.67 & 100 & 100 & 100 \\
        \#Segments & 11,376 & 6,027 & 5,349 & 1,611 & 812 & 799 \\
        \enspace\enspace \unreliable & 713 & 317 & 396 & 225 & 144 & 81 \\
        \enspace\enspace \unreliable~  pct. (\%) & 6.27 & 5.26 & 7.40 & 13.97 & 17.73 & 10.14 \\
        \#ExplainationTokens & 14,773 & 6,957 & 7,816 & 3,877 & 2,584 & 1,293 \\
        \bottomrule
    \end{tabular}
            }
    \caption{Statistics of \dataname.}
    \label{tab:data_stats}
\end{table*}

\subsection{Data Annotation}

\paragraph{Annotating Criteria for \unreliable.} We, a group of senior NLP researchers with rich Area Chairing experience, define \unreliable~review segments as follows:

\textbullet\enspace Sentences that contain factual errors or misinterpretations of the submission.

\textbullet\enspace Sentences lacking constructive feedback.

\textbullet\enspace Sentences that express overly subjective, emotional, or offensive judgments, such as ``\textit{I don't like this work because it is written like by a middle school student}.''

\textbullet\enspace Sentences that describe the downsides of the submission without supporting evidence, for example, ``\textit{This work misses some related work}.''


\textbf{Question: Why not directly use author rebuttal to infer the \unreliable~review segments?} We do not solely rely on author rebuttals for several reasons. First, author rebuttals are not always correct and may overstate contributions or include information not originally presented in the submission. Second, authors sometimes make compromises to satisfy reviewers even when the review is \unreliable. Third, author rebuttals do not address all \unreliable~details and mainly focus on the "weakness" part, while ``\unreliable'' issues can arise in other parts of the reviews.

\paragraph{Annotator Recruiting.}
Our annotator team consisted of 40 members from the NLP community, all with multiple first-authored publications in top-tier NLP venues and extensive reviewing experience. 16 have Ph.D. degrees, and 11 are university faculty members, 15 have served as area chair (AC, also called meta-reviewer in some venues) before.

\paragraph{Annotation Process.} The annotation was conducted on both human-written and LLM-generated reviews, following these steps:
i) \emph{Paper Selection}:
To ensure high-quality annotations, annotators were allowed to choose papers that aligned with their expertise and interests, ensuring their proficiency in reviewing these papers.
ii) \emph{Awareness of Review Scope}:
Our assessment focused on reviews written before the rebuttal phase, i.e., reviews based on the original submission. This decision was made to avoid the multi-turn problem and to keep the scope manageable. We did not consider extra experiments conducted during the rebuttal phase, as pre-rebuttal reviews are based on the original submission. Annotators were required to thoroughly read all reviews, meta-reviews, author rebuttals, and the original submission to ensure a comprehensive understanding of the paper and its associated reviews. 
iii) \emph{Segment-level Annotation}:
For detailed analysis, reviews were segmented by sentences, and annotators were asked to label each sentence (a) whether it is \unreliable, and (b) provide an explanation if it is. This approach allows for the identification of specific sentences that may be \unreliable, even if the overall review is of high quality. Meta-reviewers are expected to analyze individual reviews sentence by sentence.

\textbf{Question: Some reviews are generated by LLMs, how did we ensure that annotators were unaware?} 
For the annotation of LLM-generated reviews, we employed a separate group of annotators who were not informed that these reviews were LLM-generated. To prevent potential reminders for internet searches, we concealed submission information, such as "Under review as a conference paper at ICLR 2022," in the papers provided to the annotators. We acknowledge that this approach cannot guarantee complete unawareness.


\paragraph{Quality Control.} To maintain annotation quality, two annotators independently reviewed each paper's reviews without access to each other's annotations to prevent bias. Disagreements between the two annotators were resolved by a senior expert with area chair (AC) experience, who examined the conflicting annotations and resolved discrepancies by removing or rewriting the explanations for the unconvincing annotations.

\paragraph{Annotation Timeline.}Due to the time-consuming nature of high-quality annotation, each annotator was assigned one paper per week, resulting in a six-month data collection period. This ensured thorough and thoughtful annotations. We organized regular meetings to discuss any issues that arose during the annotation process.

\subsection{Data Statistics}

Table~\ref{tab:data_stats} provides the statistics for our \dataname~dataset. 
We compare from two dimensions.
First, at both the review and segment granularity, LLM-generated reviews contain more \unreliable~instances compared to human-written reviews (100\% vs. 71.57\% at the review level, and 13.97\% vs. 6.27\% at the segment level). Next, we compared ``Accepted'' and ``Rejected'', which generally correspond to ``higher-quality'' and ``lower-quality'' submissions, respectively. Notably, LLM-generated reviews demonstrated a higher frequency of \unreliable~segments for ``Accepted'' submissions than for ``Rejected'' ones, which contrasts with what we observed in human-written reviews. Drawing from our analysis in the ``Weaknesses'' part of Section \ref{sec:component}, we suggest the following explanation: human reviewers are often able to sense the overall quality of a paper. If they believe a submission is of poor quality and intend to reject it, they tend to collect more weaknesses to justify their decision, which sometimes leads to overemphasis on the paper's flaws. In contrast, LLMs lack the ability to discern paper quality and often generate superficial and non-specific criticisms \textit{equally} for both ``Accepted'' and ``Rejected'' submissions, making these review segments more likely to be inaccurate for higher-quality papers.

\subsection{Novelty of \dataname}

\begin{table}[h!]
    \small
    \centering
    \begin{tabular}{lccccc}
        \toprule
        {Dataset} & \rotatebox{90}{{PeerRead}} & \rotatebox{90}{{PRAnalyze}} & \rotatebox{90}{{Subs.PR}} & \rotatebox{90}{{DISAPERE}} & \rotatebox{90}{ReviewCrit.} \\
        \midrule
        Sentence-level & & $\checkmark$ & $\checkmark$ & $\checkmark$ & $\checkmark$ \\
        Initial submission & $\checkmark$ & & & & $\checkmark$ \\
        Highly Expert-demanding & & & & & $\checkmark$ \\
        Deficiency Labeling & & & & & $\checkmark$ \\
        Human Review & $\checkmark$ & $\checkmark$ & $\checkmark$ & $\checkmark$ & $\checkmark$ \\
        LLM review & & & & & $\checkmark$ \\
        Accepted+Rejected & $\checkmark$ & $\checkmark$ & & & $\checkmark$ \\
        
        \bottomrule
    \end{tabular}
    \caption{Comparison of \dataname~with PeerRead~\cite{kang-etal-2018-dataset}, Peer Review Analyze~\cite{ghosal2022peer}, Substantiation PeerReview~\cite{guo2023automatic} and DISAPERE~\cite{kennard-etal-2022-disapere}.}
    \label{tab:dataset_comparison}
\end{table}

As shown in Table~\ref{tab:dataset_comparison}, \dataname~ differs from previous works in several key aspects. First, \dataname~ labels review deficiencies at the sentence level, demanding highly experienced annotators. Second, annotators must read the initial submission, meta-reviews, all reviews, and rebuttals before annotating, unlike previous works that require reading reviews and, at most, rebuttals. These differences make \dataname~ the only dataset suitable for benchmarking LLMs as responsible meta-reviewers, offering a comprehensive evaluation of review quality. Additionally, \dataname~ includes expert-annotated LLM-generated reviews, enabling direct comparison between human and LLM-generated reviews at a granular level. These unique features distinguish \dataname~ and open new research opportunities in AI for peer review.

\section{Experiments}
 We present experimental results and analysis in two threads: \llmasreviewers~ (Section~\ref{sec:llm_as_reviewer}), and \llmasmeta~(Section \ref{sec:llm_as_metareviewer}).

\subsection{\llmasreviewers~(i.e., Human-written reviews vs. LLM-generated reviews)} 

\label{sec:llm_as_reviewer}

In this section, we compare LLM-generated reviews with human-written reviews: i) by the fine-grained error types if the review segments are annotated \unreliable, ii) by fine-grained analysis for each component (summary, strengths, weakness, writing, and recommendation score), iii) by considering review diversity.

\subsubsection{Error type analysis for deficiency}\label{sec:errortype}

Besides the coarse-grained ``\unreliable'' label, our annotation team classify the expert-annotated \unreliable~segments into  23 fine-grained error types (full list and their explanations  in Table~\ref{tab:error_types_explanations}, Appendix~\ref{appendix_error_types}). Table~\ref{tab:error_types} (Appendix~\ref{appendix_error_types}) report the percentage of each error type for both human-written and LLM-generated reviews. Table~\ref{tab_top3_errors} shows the comparison of the top-3 most frequent error types between human and LLM reviews. 

\begin{table}[t]
\centering
\resizebox{0.9\linewidth}{!}{
\begin{tabular}{@{}lrr@{}}
\toprule
\textbf{Error Type}       & {\textbf{Human (\%)}} & \textbf{LLM (\%)} \\ 
\midrule
\rowcolor{gray!30} 
\multicolumn{3}{c}{\textit{Human top-3}} \\
Misunderstanding          & 22.86        & 9.87    \\
Neglect                   & 19.64        & 5.83      \\
Inexpert Statement        & 18.23        & 6.73     \\
\addlinespace
\rowcolor{gray!30} 
\multicolumn{3}{c}{\textit{LLM top-3}}  \\
Out-of-scope              & 4.35         & 30.49     \\
Misunderstanding          & 22.86        & 9.87    \\
Superficial Review        & 2.66         & 9.42      \\
\bottomrule
\end{tabular}
}
\caption{Comparing top-3 error types between human-written and LLM-generated reviews.}
\label{tab_top3_errors}
\end{table}

From Table ~\ref{tab_top3_errors}, a major reason for \unreliable~ reviews from human reviewers is misunderstanding the paper submission and raising unnecessary concerns by neglecting information already stated. This suggests a lack of patience during the reviewing process. Another significant error is making inexpert critiques or statements due to insufficient domain knowledge, potentially from unqualified reviewers being involved due to the increasing number of submissions to AI/NLP conferences and the need to recruit more reviewers.


Compared to humans, LLMs are more likely to suggest out-of-scope experiments or analyses. They make significantly fewer "Inexpert Statement" errors. Based on our observations, this is because their reviews are usually paper-unspecific and superficial, avoiding expert-level mistakes. Additionally, LLM-generated reviews do not exhibit errors like "Missing Reference," "Invalid Reference," and "Concurrent Work" since they do not point to specific works or provide references.

\subsubsection{Fine-grained review analysis}\label{sec:component}

\paragraph{``Summary'' part.} The Summary section in LLM-generated reviews exhibits relatively better quality compared to other aspects. Our annotators identified only 1.35\% of segments as "Inaccurate Summary" among all LLM \unreliable~segments, which constitutes 0.19\% of all LLM-generated segments. In comparison, 5.75\% of segments were identified as "Inaccurate Summary" among all \unreliable~segments in human-written reviews, accounting for 0.36\% of all human-written review segments. This is nearly twice the percentage found in LLM-generated summaries.
Moreover, error types such as ``Summary Too Short'' and ``Copy-pasted Summary'', which are present in human reviews, were not observed in LLM-generated reviews, suggesting that LLMs are capable of generating summaries of satisfying quality and avoid directly copying content from the paper.

\paragraph{``Strengths'' part.} LLMs tend to accept authors' claims in submissions without much critical evaluation. Our analysis reveals that among all segments in the Strengths section of LLM-generated reviews, 53.2\% are simply rephrased from the submission, while the remaining segments are mostly inferred from the introduction and abstract, where authors typically highlight their contributions.

To further investigate, we used \dataname~ to compare human-written reviews assessed by annotators and LLM-generated reviews for the same papers. For accepted papers, 34.5\% of the Strength segments generated by LLMs were questioned by human experts in their corresponding human-written reviews. For rejected papers, this rose to 51.9\%.


These findings suggest that LLMs often accept authors' claims without thorough verification, treating strengths as a text summarization task. In contrast, human reviewers scrutinize the claimed strengths and provide their expert opinions on the validity and significance of the contributions.

\paragraph{``Weaknesses'' part.}


The most dominant type of \unreliable~in LLM reviews is ``Out-of-scope'', accounting for 30.49\% of all \unreliable~segments in LLM-generated reviews (see Table~\ref{tab_top3_errors}). LLMs often highlight weaknesses such as the need for more experiments, lack of generalizability, additional tasks, more analysis, evaluation on languages beyond English, etc. While occasionally relevant, these suggestions often fall outside the paper's scope and shouldn't be considered weaknesses.

Moreover, the suggestions provided by LLMs in the Weaknesses section tend to be paper-unspecific and superficial (e.g, \textit{The paper's focus on pre-trained models might limit its applicability to domains where such models are not available or suitable.}), making them applicable to most NLP papers without offering actionable insights to either authors or area chairs. This lack of specificity and depth in the critiques highlights the limitations of LLMs in providing meaningful and constructive feedback on the weaknesses of a paper.

These findings underscore the importance of human expertise in identifying and articulating the most relevant and significant weaknesses of a paper. While LLMs can generate a list of potential limitations, they often struggle to contextualize these weaknesses within the scope and objectives of the paper, leading to \unreliable~segments that may not be helpful to authors or area chairs.

\paragraph{``Writing'' part.}
Our analysis suggests that LLMs may lack the ability to accurately judge the writing quality of a paper submission. In all LLM-generated reviews, LLMs consistently praise the writing of the papers, stating that they are well-written and easy to follow. However, among the papers used for generating LLM reviews, 15\% of the papers had both the meta-reviewer and human reviewers agree that the writing was unclear and difficult to follow. Despite this consensus among human experts, the LLMs still provided positive feedback on the writing quality of these papers, failing to accurately assess the writing quality.

\paragraph{``Recommendation Score'' part.}


In addition to generating reviews, we asked LLMs to rate each paper on a scale of 1-10, matching the ICLR and NeurIPS system, for directly comparison with human reviewers. Experiment shows that LLMs tend to give high scores to all submissions, regardless of quality or acceptance status, with averages of 7.43 for accepted and 7.47 for rejected papers. In contrast, human reviewers scored averages of 6.41 for accepted and 4.81 for rejected submissions. Thus, LLMs do not align with the human reviewers in discerning paper quality based on their internal scoring mechanism.




\subsubsection{Review Diversity}
\label{sec_reviewr_diversity}
Given three LLMs and $m$ papers, we can get a matrix of LLM-generated reviews of size $3\times m$. We perform quantitative analysis i) horizontally to measure the  ``intra-LLM review specificity'', and ii) vertically as the assessment of ``inter-LLM review complementarity''.

\paragraph{Intra-LLM Review Specificity.} 

In the real world, we hope the review for each paper is specific to this paper. Then the paper-specific review diversity should discourage two cases: i) one review has too many repeat of certain segment; ii) a review segment appear in too many papers. We get inspiration from the classic TF-IDF to define a new segment-level diversity metric, named \textbf{\newmetric}:
\begin{equation}
\scriptsize
\intrametric=\frac{1}{m} \sum_{j=1}^{m} \left( \frac{1}{n_j} \sum_{i=1}^{n_j} \log \left( \frac{n_j}{O_i^j} \right) \times \log \left( \frac{m}{R_i^j} \right) \right),
\end{equation}
where \(n_j\) is the number of segments in review \(j\), \(O_i^j\) is the ``soft'' occurrence of segment \(s_i^j\) in review \(j\), \(R_i^j\) is the ``soft'' number of reviews containing segment \(s_i^j\). \(O_i^j\) is computed as follows:
\begin{equation}
\small
O_i^j = \sum_{k=1}^{n_j} \mathbb{I}(\text{sim}(s_i^j, s_k^j) \geq t) \cdot \text{sim}(s_i^j, s_k^j),
\end{equation}
where \(s_i^j\) and \(s_k^j\) are the \(i\)-th and \(k\)-th segments in review \(j\), respectively. \(O_i^j\) is calculated by summing the similarity scores between segment \(s_i^j\) and all other segments \(s_k^j\) in the same review \(j\) that exceed a predefined similarity threshold \(t\). \(R_i^j\) is  defined as follows:
\begin{equation}
\small
R_i^j = \sum_{l=1}^{m} \mathbb{I}\left(\max_{p} \text{sim}(s_i^j, s_p^l) \geq t\right) \cdot \max_{p} \text{sim}(s_i^j, s_p^l),
\end{equation}
where \(s_p^l\) is any segment in review \(l\). \(R_i^j\) is computed by summing the maximum similarity scores between segment \(s_i^j\) and segments in each review \(l\) that exceed the threshold \(t\). In our experiments, we use SentenceBERT~\cite{DBLP:conf/emnlp/ReimersG19} to calculate the similarity between segments. Implementation details can be found in Appendix~\ref{appendix_simscore}.

In summary, \intrametric~measures the specificity of reviews generated by a single LLM across different papers. A lower \intrametric~score means LLM tends to generate repetitive or similar segments across reviews, while a higher score suggests more diverse and unique content in the generated reviews.

\begin{figure}
    \centering
    \includegraphics[width=0.48\textwidth]{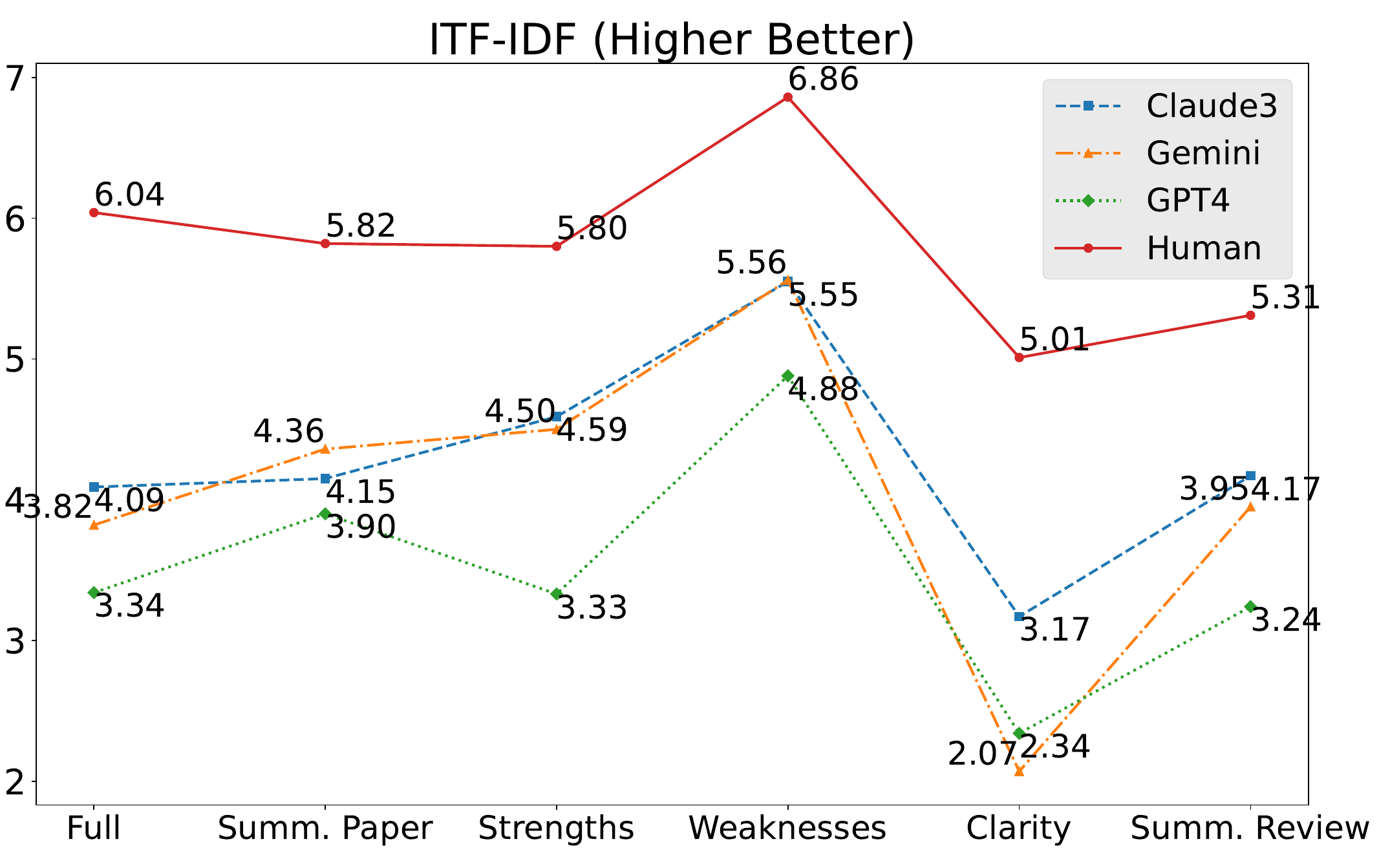}
    \caption{Specificity of reviews: LLM vs. Human.}
    \label{fig:itf_idf}
\end{figure}

Figure~\ref{fig:itf_idf} shows the Intra-LLM paper-oriented specificity on different review components such as strengths, weaknesses, etc. We set threshold $t$ as 0.5 because our initial observation suggests that segments with a similarity higher than this threshold have a similar meaning. We also report the evaluations under different $t$ values in Table~\ref{tab:intrametric_diff_t} (Appendix~\ref{appendix_threshold}). For human-written reviews, we randomly sample one review from each paper and calculate \newmetric. We repeat this process five times and use the average score. 

For \intrametric, from the full review perspective, human reviews score the highest (6.04), followed by Claude Opus (4.09), Gemini (3.82), and GPT-4 (3.34). The scores are relatively consistent across different sections, but GPT-4 tends to have the lowest scores, suggesting more repetitive segments compared to other LLMs. Human reviews maintain high diversity across all sections. LLMs exhibit a sharp diversity drop in the ``Clarity'' section. This aligns with our observation in Section~\ref{sec:component} that LLMs praise the writing quality of all papers. 

\paragraph{Inter-LLM Review Complementarity.}

We examine whether different LLMs tend to write complementary reviews for the same paper, which is a pairwise concept. We first compute the BERTScore~\cite{BERTScore} for each pair of reviews generated by the three LLMs (GPT-4, Claude Opus, and Gemini 1.5) for the same paper. We then average these scores across all papers to obtain an overall measure of Inter-LLM review diversity. 

\begin{figure}
    \centering
    \includegraphics[width=0.8\linewidth]{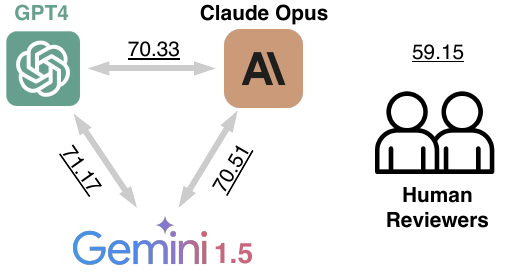}
    \caption{Inter-LLM vs. inter-human review similarities.}
    \label{fig:inter_score}
\end{figure}

Figure~\ref{fig:inter_score} shows the pairwise BERTScores for reviews on the same paper generated by GPT-4, Claude Opus, and Gemini 1.5. It also presents the BERTScores for reviews of the same paper conducted by human reviewers. The BERTScores between different LLM pairs are similar and high, ranging from 70.33 to 71.17. In comparison, the BERTScore between human reviewers is 59.15, which is noticeably lower than the scores between the LLMs. This indicates that human reviewers tend to produce more diverse reviews compared to the LLMs. In addition, this finding implies that the use of multiple LLMs may not necessarily lead to a significant increase in the diversity of perspectives and insights in the review process.

\subsection{\llmasmeta}
\label{sec:llm_as_metareviewer}

\begin{table*}[h!]
    \centering
    \setlength{\tabcolsep}{3.5pt}
    \begin{tabular}{@{}lrrrr@{}}
        \toprule
        \multirow{2}{*}{Model} & \multicolumn{4}{c}{Precision / Recall / F1} \\
        \cmidrule(lr){2-5}
        & \multicolumn{1}{c}{\promptall} & \promptneg & \multicolumn{1}{c}{Both ``No''} & \multicolumn{1}{c}{Either ``No''} \\
        \midrule
        GPT-4 & 14.91 / 34.49 / 18.38 & 17.18 / 34.59 / 20.30 & 18.71 / 21.40 / 16.85 & 14.72 / 47.68 / \underline{20.66} \\
        Claude Opus & 16.86 / 34.26 / 20.35 & 17.69 / 26.61 / 18.71 & 17.14 / 18.70 / 15.78 & 16.94 / 42.12 / \textbf{\underline{21.99}} \\
        Gemini 1.5 & 16.58 / 34.13 / 19.76 & 14.71 / 43.60 / 19.72 & 17.01 / 27.05 / 18.28 & 14.46 / 50.37 / \underline{20.34} \\
        Llama3-8B & 7.73 / 45.95 / 12.22 & 11.47 / 30.29 / \underline{14.88} & 11.37 / 21.27 / 12.46 & 8.19 / 53.61 / 13.35 \\
        Llama3-70B & 13.63 / 42.49 / 18.19 & 13.95 / 31.16 / 17.46 & 16.16 / 23.51 / 16.67 & 12.46 / 50.02 / \underline{18.43} \\
        Qwen2-72B & 9.97 / 26.60 / 12.96 & 11.35 / 34.61 / 14.64 & 9.07 / 15.13 / 9.62 & 10.49 / 43.00 / \underline{15.16} \\
        \bottomrule
    \end{tabular}
    \caption{Performance of LLMs as meta-reviewers on our \dataname~dataset. The best F1 score among different prompt methods for a single model is \underline{underlined}. The best F1 score across all models is also \textbf{\underline{bold}}.}
    \label{tab:llm_meta_reviewer_prf}
\end{table*}

As an area chair, one should assess the quality of individual reviews using their own expertise. This task is highly knowledge-intensive and requires deep understandings of the research domain. Our \dataname~provides segment-level annotation on if each segment is deficient and why. This section evaluates if prompting popular LLMs (both closed- and open-source) can solve this problem. For closed-source models, we assess GPT4~\cite{openai2023gpt4}, Claude Opus~\cite{claude}, and Gemini1.5~\cite{google2023gemini}. For open-source models, we evaluate Llama3-8B and -70B~\cite{llama3modelcard} and Qwen2-72B~\cite{qwen}.

To mitigate the impact of prompt-specific performance, we employ two prompting strategies: 1)
\textbf{\promptall}: Given everything necessary including a list of indexed review segments, require the LLM to output a list of triples like (id, \unreliable~or not, explanation);
2) \textbf{\promptneg}: Given everything necessary including a list of indexed review segments, require the LLM to output a list of tuples, (id,  explanation),  when it believes the ``id'' corresponds to an \unreliable~segment.  
The detailed prompt templates are in Table~\ref{tab:assess_all} and~\ref {tab:identify_unreliable} (Appendix~\ref{appendix_prompt_templates}).

To enhance evaluation robustness, we ensemble the results obtained from the two prompting strategies using two methods: i) \textbf{Both ``No''}: If both prompts classify a segment as \unreliable, we consider it to be \unreliable; ii) \textbf{Either ``No''}: If either of the prompts labels a segment as \unreliable, we consider it to be \unreliable.

\paragraph{How well can LLMs identify the \unreliable~segments experts discovered?}

Metric: we compute the F1 on each paper then average across papers.
Table~\ref{tab:llm_meta_reviewer_prf} presents the evaluation results.

Closed-source models (GPT-4, Claude Opus, and Gemini 1.5) generally outperform open-source models (Llama3-8B and 70B, Qwen2-72B) in F1 score. Claude Opus achieves the highest F1 scores, with GPT-4 and Gemini 1.5 performing slightly worse. Notably, ``recall'' scores are consistently higher than precision scores across all LLMs and prompting strategies, suggesting that LLMs tend to incorrectly identify segments as \unreliable.

Despite the superior performance of the closed-source models, their F1 scores remain relatively low even with different prompt strategies, highlighting the challenges LLMs face in such expertise-intensive tasks and emphasizing the importance of human expertise in the meta-reviewing process.

\paragraph{Can LLMs correctly explain their ``\unreliable'' judgment?}
When LLM's label \unreliable~is correct, we calculate ROUGE~\cite{lin2004rouge} and BERTScores between its explanations and our expert's explanations. Table~\ref{tab:explanation_score} reports evaluation results for the \promptneg~prompt. The full scores for both prompt strategies and their ensembles are in Table~\ref{tab:explanation_score_full_1} and~\ref{tab:explanation_score_full_2} in Appendix~\ref{appendix_explanation_score_full}.

\begin{table}[t]
    \centering
    \resizebox{0.85\linewidth}{!}{
    \begin{tabular}{@{}lc@{}}
        \toprule
        {Model} & {ROUGE-1/2/L/BERTScore} \\
       \midrule
        {GPT-4} & 17.13 / 2.71 / 14.64 / 55.63 \\ 
        {Claude Opus} &  \textbf{20.18} / \textbf{3.69} / \textbf{17.52} / \textbf{57.28}  \\ 
        {Gemini 1.5} &  18.47 / 2.98 / 16.38 / 56.46  \\ 
        {Llama3-8B} & 16.49 / 2.22 / 13.65 / 55.23 \\ 
        {Llama3-70B} & 15.94 / 1.95 / 13.78 / 57.09 \\ 
        {Qwen2-72B} & 17.07 / 3.00 / 14.69 / 56.88 \\ 
        \bottomrule

    \end{tabular}
    }
    \caption{Evaluation of LLMs' explanations for correctly identified \unreliable~segments.}
    \label{tab:explanation_score}
\end{table}

The results in Table~\ref{tab:explanation_score} show that overall scores for all LLMs are relatively low, indicating they can identify some \unreliable~segments but struggle to articulate their reasoning.
Among the LLMs, Claude Opus achieves the highest scores across all metrics, suggesting its explanations align best with human annotators. 
Claude Opus also excels in identifying \unreliable~segments, as shown previously. GPT-4 and Gemini 1.5 show similar performance to Claude Opus. The open-source models, Llama3 (8B and 70B) and Qwen2-72B, generally score lower than the closed-source models.

\paragraph{Which \unreliable~types are challenging for LLMs to identify?}
To investigate which types of \unreliable~are more challenging for LLMs to detect, we check for each \unreliable~type how many can be successfully identified by LLMs. We focus on three closed-source LLMs: GPT-4, Claude Opus, and Gemini 1.5. 

Table~\ref{tab:llm_identified_errors} (in Appendix~\ref{appendix_error_types_detected}) presents the number and percentage of segments identified in each \unreliable~type by the LLMs. We observe that six types of \unreliable~have a significantly lower percentage compared to the average recall of GPT-4 (47.68\%), Claude Opus (42.12\%), and Gemini 1.5 (50.37\%), suggesting that these types of \unreliable~ are particularly difficult for LLMs to detect: \texttt{Inaccurate Summary}, \texttt{Writing}, \texttt{Superficial Review}, \texttt{Experiment}, \texttt{Contradiction} and \texttt{Unstated Statement} 

These findings align with our observations in Sections~\ref{sec:component}\&\ref{sec:llm_as_reviewer}, where we assessed LLMs as reviewers. For example, LLMs struggle to accurately judge the paper writing quality  submission and tend to provide superficial reviews, often failing to offer constructive suggestions on experiments. Moreover, LLMs are more prone to generating contradictory claims in their reviews and making claims that the authors never stated in the submission, indicating a tendency towards hallucination. Additionally, although LLMs can generate paper summaries with fewer errors, they may fail to capture nuanced aspects of the paper, leading to their inability to identify inaccurate summary errors.
\section{Conclusion}

This work studied the potential of \topic, focusing on their roles as reviewers and meta-reviewers. We created \dataname~, containing both human-written and LLM-generated reviews, with detailed deficiency annotations and explanations. Our analysis reveals that while LLMs can generate reviews, they often produce \unreliable~and paper-unspecific segments, lacking the diversity and constructive feedbacks. Additionally, even state-of-the-art LLMs struggle to assess review deficiencies effectively. These findings highlight the current limitations of LLMs in automating the peer review process.


\section*{Limitations}

While our work provides valuable insights into the potential of LLMs in the peer review process, there are some limitations. During the evaluation of LLMs, \dataname~primarily focuses on the textual information from the submissions and does not include figures, tables, or other visual elements. Incorporating these additional components could provide a more comprehensive assessment of LLMs' capabilities in the peer review process. Additionally, the dataset is currently limited to the NLP domain. It would be interesting to explore the performance of LLMs in other research areas. Expanding the dataset to include papers from various domains could help assess the generalizability of our findings and identify potential domain-specific challenges. Furthermore, our work focuses on the pre-rebuttal phase of the peer review process, assessing reviews based on the original submission. Incorporating the multi-turn aspect of peer review, including author rebuttals and post-rebuttal reviews, could offer a more comprehensive understanding of LLMs' capabilities in the entire review process. 

\section*{Ethical Considerations}
In this study, we carefully considered the ethical implications of using LLMs to assist in the review process. We acknowledge potential risks, including bias, lack of accountability, and the possibility of undermining the integrity of scientific evaluations. Importantly, this work does not advocate for the use of LLMs in paper (meta-)reviewing. Rather, our study underscores that LLMs are currently insufficient to replace human reviewers, especially in tasks that require expert judgment and nuanced understanding.

\section*{Acknowledgments}
The authors appreciate the reviewers for their insightful comments and suggestions.
This work is supported in part by NSF under grants III-2106758, and POSE-2346158.

\bibliography{custom}

\appendix

\section{Experiment Details}

\label{appendix_implementation_details}

\subsection{BERTScore}
During the evaluation of LLMs' explanations for \unreliable~segments (Table~\ref{tab:explanation_score}, \ref{tab:explanation_score_full_1}, and \ref{tab:explanation_score_full_2}), we use \texttt{microsoft/deberta-xlarge-mnli} as the base model for computing BERTScore~\cite{BERTScore}, as officially suggested\footnote{\url{https://github.com/Tiiiger/bert\_score}}.

In the experiment of computing inter-LLM review complementarity (Section~\ref{sec:llm_as_reviewer}), we use \texttt{facebook/bart-large-mnli} as the base model for BERTScore. This is because \texttt{microsoft/deberta-xlarge-mnli} only supports input sequences up to 512 tokens, while some full reviews exceed this limit. In contrast, \texttt{facebook/bart-large-mnli} has a context size of 1024 tokens, making it suitable for processing longer reviews.

\subsection{Similarity Score in \newmetric}

\label{appendix_simscore}

We use SentenceBERT~\cite{DBLP:conf/emnlp/ReimersG19} to calculate the similarity in \newmetric. We adopt the \texttt{all-MiniLM-L6-v2} pretrained model because it is fast and still offers good quality~\footnote{\url{https://sbert.net/docs/sentence\_transformer/pretrained\_models.html}}. In practice, the similarity in our \newmetric~can be computed using any sentence similarity model.

\subsection{LLM Inference Details}

\paragraph{Closed-source LLMs.} We experiment with the following models and their corresponding API endpoints: GPT-4 (\texttt{gpt-4-turbo}), Gemini 1.5 (\texttt{gemini-1.5-flash-latest}), and Claude 3 (\texttt{claude-3-opus-20240229}).

\paragraph{Open-source LLMs.} We experiment with the following models: Llama3-8B (\texttt{Meta-Llama-3-8B-Instruct}), Llama3-70B (\texttt{Meta-Llama-3-70B-Instruct}), and Qwen2-72B (\texttt{Qwen/Qwen2-7B-Instruct}).

For GPT-4, Claude 3, Gemini 1.5, and Qwen2-72B, we input the full prompt as shown in Table~\ref{tab:assess_all} and \ref{tab:identify_unreliable}, which contains the complete instruction, paper title, full paper body text, and review text.

However, for Llama3-8B and Llama3-70B, the maximum supported context length is limited to 8k tokens\footnote{\url{https://github.com/meta-llama/llama3/blob/main/MODEL\_CARD.md}}. To accommodate this constraint, we truncate the full paper body text while keeping the other components of the prompt intact. This is because the other components, such as the instruction, paper title, and review text, are crucial for the evaluation and cannot be truncated.

\section{Influence of Different Thresholds in \intrametric}
\label{appendix_threshold}

Table~\ref{tab:intrametric_diff_t} shows the impact of varying the similarity threshold $t$ on the \intrametric~scores on full reviews. The performance rank remains the same across different $t$ values. 

\begin{table}[h!]
    \small
    \centering
    \begin{tabular}{lcccc}
        \toprule
        \textbf{Model} & \textbf{$t$ = 0} & \textbf{$t$ = 0.5} & \textbf{$t$ = 0.7} & \textbf{$t$ = 0.99} \\
        \midrule
        GPT4 & 0.77 & 3.37 & 6.46 & 8.42 \\
        Claude3 & 0.87 & 4.09 & 7.61 & 9.32 \\
        Gemini & 0.81 & 3.82 & 6.67 & 8.57 \\
        Human & 1.22 & 6.04 & 8.45 & 9.50 \\
        \bottomrule
    \end{tabular}
    \caption{\intrametric~ under different $t$ values. The rank remains the same across different $t$ values.}
    \label{tab:intrametric_diff_t}
\end{table}

\section{Error Types Detected by LLMs}
\label{appendix_error_types_detected}

 Table~\ref{tab:llm_identified_errors} provide a statistics of the error types that LLMs successfully identify in the human-written reviews. We report the number and percentage of segments detected by each LLM for each error type.

\begin{table*}[h!]
    \centering
    \begin{tabular}{@{}lccc@{}}
        \toprule
        \multirow{2}{*}{Error Type} & \multirow{2}{*}{\dataname} & \multicolumn{2}{c}{GPT4 / Claude Opus / Gemini 1.5} \\
        \cmidrule(lr){3-4}
         &  & Identified & Percentage \\
        \midrule
        Out-of-scope & 31 & 14 / 21 / 25 & 45.2 / 67.7 / 80.6\% \\
        \rowcolor{red!30} Inaccurate Summary & 41 & 7 / 2 / 4 & 17.1 / 4.9 / 9.8\% \\
        Neglect & 140 & 75 / 100 / 122 & 53.6 / 71.4 / 87.1\% \\
        Inexpert Statement & 130 & 68 / 80 / 100 & 52.3 / 61.5 / 76.9\% \\
        Misunderstanding & 163 & 77 / 111 / 120 & 47.2 / 68.1 / 73.6\% \\
        Vague Critique & 66 & 39 / 52 / 57 & 59.1 / 78.8 / 86.4\% \\
        Misinterpret Novelty & 27 & 19 / 23 / 22 & 70.4 / 85.2 / 81.5\% \\
        Misplaced Attributes & 7 & 4 / 3 / 5 & 57.1 / 42.9 / 71.4\% \\
        \rowcolor{red!30} Writing & 20 & 2 / 2 / 4 & 10.0 / 10.0 / 20.0\% \\
        \rowcolor{red!30} Superficial Review & 19 & 2 / 2 / 3 & 10.5 / 10.5 / 15.8\% \\
        Invalid Criticism & 20 & 11 / 12 / 16 & 55.0 / 60.0 / 80.0\% \\
        Invalid Reference & 3 & 2 / 1 / 2 & 66.7 / 33.3 / 66.7\% \\
        Subjective & 8 & 5 / 7 / 6 & 62.5 / 87.5 / 75.0\% \\
        Missing Reference & 9 & 6 / 7 / 7 & 66.7 / 77.8 / 77.8\% \\
        \rowcolor{red!30} Experiment & 13 & 2 / 3 / 3 & 15.4 / 23.1 / 23.1\% \\
        \rowcolor{red!30} Contradiction & 5 & 1 / 1 / 1 & 20.0 / 20.0 / 20.0\% \\
        Summary Too Short & 2 & 1 / 0 / 1 & 50.0 / 0.0 / 50.0\% \\
        Typo & 2 & 1 / 2 / 2 & 50.0 / 100.0 / 100.0\% \\
        Concurrent Work & 1 & 1 / 1 / 1 & 100.0 / 100.0 / 100.0\% \\
        \rowcolor{red!30} Unstated Statement & 2 & 0 / 0 / 0 & 0.0 / 0.0 / 0.0\% \\
        Copy-pasted Summary & 2 & 0 / 0 / 1 & 0.0 / 0.0 / 50.0\% \\
        Misunderstanding Submission Rule & 2 & 1 / 2 / 2 & 50.0 / 100.0 / 100.0\% \\
        \bottomrule
    \end{tabular}
    \caption{Comparison of GPT-4, Claude, and Gemini in identifying \unreliable~segments. Red-colored types have a significantly lower percentage compared to the average recall of LLMs.}
    \label{tab:llm_identified_errors}
\end{table*}

\section{\unreliable~Segment Error Types}
\label{appendix_error_types}

Table~\ref{tab:error_types_explanations} present a comprehensive list of the error types used to categorize the \unreliable~segments in the reviews. Each error type is accompanied by an explanation defined by our annotation team. We also report the percentage of each error type for both human-written and LLM-generated reviews in Table~\ref{tab:error_types}.

\begin{table*}[h]
    \centering
    \begin{tabular}{lcccccc}
        \toprule
        \textbf{Error Types} & \multicolumn{3}{c}{\textbf{Human Review}} & \multicolumn{3}{c}{\textbf{LLM Review}} \\
        \cmidrule(lr){2-4} \cmidrule(lr){5-7}
        & \textbf{All} & \textbf{Acc.} & \textbf{Rej.} & \textbf{All} & \textbf{Acc.} & \textbf{Rej.} \\
        \midrule
        Out-of-scope & 4.35\% & 5.05\% & 3.79\% & 30.49\% & 33.80\% & 24.69\% \\
        Inaccurate Summary & 5.75\% & 8.52\% & 3.54\% & 1.35\% & 0.70\% & 2.47\% \\
        Neglect & 19.64\% & 24.29\% & 15.91\% & 5.83\% & 4.23\% & 8.64\% \\
        Inexpert Statement & 18.23\% & 16.72\% & 19.44\% & 6.73\% & 4.23\% & 11.11\% \\
        Misunderstanding & 22.86\% & 17.35\% & 27.27\% & 9.87\% & 11.97\% & 6.17\% \\
        Vague Critique & 9.26\% & 5.99\% & 11.87\% & 7.17\% & 4.23\% & 12.35\% \\
        Misinterpret Novelty & 3.79\% & 6.94\% & 1.26\% & 2.24\% & 2.82\% & 1.23\% \\
        Misplaced attributes & 0.98\% & 0.95\% & 1.01\% & - & - & - \\
        Writing & 2.81\% & 2.52\% & 3.03\% & 4.48\% & 4.23\% & 4.94\% \\
        Superficial Review & 2.66\% & 3.15\% & 2.27\% & 9.42\% & 7.04\% & 13.58\% \\
        Invalid Criticism & 2.81\% & 2.84\% & 2.78\% & - & - & - \\
        Invalid Reference & 0.42\% & 0.32\% & 0.51\% & - & - & - \\
        Subjective & 1.12\% & 1.89\% & 0.51\% & - & - & - \\
        Missing Reference & 1.26\% & 0.63\% & 1.77\% & - & - & - \\
        Experiment & 1.82\% & 1.89\% & 1.77\% & 1.35\% & 1.41\% & 1.23\% \\
        Contradiction & 0.70\% & - & 1.26\% & 8.52\% & 10.56\% & 4.94\% \\
        Summary Too Short & 0.28\% & - & 0.51\% & - & - & - \\
        Typo & 0.28\% & - & 0.51\% & - & - & - \\
        Concurrent work & 0.14\% & - & 0.25\% & - & - & - \\
        Unstated statement & 0.28\% & 0.63\% & - & 7.62\% & 7.75\% & 7.41\% \\
        Copy-pasted Summary & 0.28\% & - & 0.51\% & - & - & - \\
        Misunderstanding Submission Rule & 0.28\% & 0.32\% & 0.25\% & - & - & - \\
        Duplication & - & - & - & 4.93\% & 7.04\% & 1.23\% \\
        \bottomrule
    \end{tabular}
    \caption{Percentage fo error types in Human-written and LLM-generated reivews amaong all \unreliable~segments.}
    \label{tab:error_types}
\end{table*}

\begin{table*}[ht]
\centering

\begin{tabular}{p{0.25\linewidth}p{0.7\linewidth}}
\toprule
\textbf{Error Type} & \textbf{Explanation} \\
\midrule
Misunderstanding & The reviewer misinterprets claims or ideas presented in the paper, leading to inaccurate or irrelevant comments. \\
Neglect & The reviewer overlooks important details explicitly stated in the paper, resulting in unwarranted questions or critiques. \\
Vague Critique & The review lacks specificity, claiming missing components without clearly identifying what is missing. \\
Inaccurate Summary & The summary in the review misrepresents the main content or contributions of the paper.  \\
Out-of-scope & The reviewer suggests additional methods, experiments, or analyses that are beyond the intended scope of the paper. \\
Misunderstanding of the Submission Rule & The reviewer believes the submission format violates conference rules, but this is not actually the case. \\
Subjective & The review makes assertions about the paper's clarity or quality without providing sufficient justification or evidence. \\
Invalid Criticism & The reviewer's criticism is considered invalid, especially when suggesting impractical experiments or trivializing results. \\
Misinterpret Novelty & The reviewer questions the novelty of the work without substantiating their claims with relevant references \\
Superficial Review &  The reviewer appears to have only skimmed the paper, providing generic or unsupported comments about the presence or absence of weaknesses.  \\
Writing & Discrepancies arise when the reviewer praises the writing, while our annotator suggests it needs more clarity or explicitness. \\
Inexpert Statement & The reviewer exhibits a lack of domain knowledge, leading to unnecessary or irrelevant concerns. \\
Missing Reference & The reviewer proposes alternative frameworks or methods without providing justification or citing relevant references \\
Experiment & Conflicting opinions about the design of experiments; the reviewer praises them while our annotator suggests adding more baselines or tests. \\
Misplaced attributes & Strengths are incorrectly listed as weaknesses or vice versa. \\
Invalid Reference & The reviewer cites non-peer-reviewed sources or blogs, which is not appropriate for academic validation. \\
Unstated statement & Statements made in the review are not supported by content in the paper. \\
Summary Too Short & The provided summary is excessively brief, offering little to no insight into the actual content of the paper.  \\
Contradiction & The reviewer contradicts themselves within the review, such as criticizing the paper's experiments while later stating that the experiments are comprehensive. \\
Typo & The review contains typographical errors that may affect clarity or understanding. \\
Copy-pasted Summary & The summary is directly copied from the submission. \\
Concurrent work & The reviewer requests comparisons with work conducted concurrently, which may not have been considered by the authors. \\
Duplication & The review segment is a repetition or duplication of a previous segment within the same review. \\
\bottomrule
\end{tabular}
\caption{Error types in paper reviews.}
\label{tab:error_types_explanations}
\end{table*}

\section{Explanation Score Across Different Prompts}
\label{appendix_explanation_score_full}

This section compares the performance of LLMs in generating explanations for the correctly identified \unreliable~segments across different prompting strategies. We report the ROUGE and BERTScore values for each LLM and prompt combination in Table~\ref{tab:explanation_score_full_1} and~\ref{tab:explanation_score_full_2}.

\begin{table*}[h!]
    \centering
    
    \begin{tabular}{@{}lcc@{}}
        \toprule
        \multirow{2}{*}{Model} & \multicolumn{2}{c}{ROUGE-1 / 2 / L / BERTScore} \\
        \cmidrule(lr){2-3}
        & \promptall & \promptneg  \\
       \midrule
        {GPT-4} & 16.12 / 2.05 / 13.58 / 56.87 & 17.13 / 2.71 / 14.64 / 55.63 \\ 
        {Claude Opus} & 18.54 / 3.03 / 16.03 / 58.44 & 20.18 / 3.69 / 17.52 / 57.28  \\ 
        {Gemini 1.5} & 19.40 / 2.99 / 17.14 / 58.10 & 18.47 / 2.98 / 16.38 / 56.46  \\ 
        {Llama3-8B} & 15.97 / 1.74 / 14.14 / 56.23 & 16.49 / 2.22 / 13.65 / 55.23 \\ 
        {Llama3-70B} & 15.03 / 2.25 / 13.04 / 58.19 & 15.94 / 1.95 / 13.78 / 57.09 \\ 
        {Qwen2-72B} & 14.49 / 2.27 / 12.86 / 56.66 & 17.07 / 3.00 / 14.69 / 56.88 \\ 
        
        \bottomrule

    \end{tabular}
    \caption{Evaluation of LLMs' explanations for correctly identified \unreliable~segments with \promptall~and \promptneg~prompt methods. }
    \label{tab:explanation_score_full_1}
\end{table*}

\begin{table*}[h!]
    \centering
    
    \begin{tabular}{@{}lcc@{}}
        \toprule
        \multirow{2}{*}{Model} & \multicolumn{2}{c}{ROUGE-1 / 2 / L / BERTScore} \\
        \cmidrule(lr){2-3}
        & Both "No" & Either "No" \\
       \midrule
        {GPT-4} & 16.79 / 2.46 / 14.16 / 56.21 & 16.61 / 2.36 / 14.09 / 56.25 \\ 
        {Claude Opus} & 19.82 / 3.63 / 17.23 / 58.00 & 19.24 / 3.31 / 16.66 / 57.95 \\ 
        {Gemini 1.5} & 19.25 / 3.08 / 17.12 / 57.42 & 18.88 / 2.99 / 16.72 / 57.17 \\ 
        {Llama3-8B} & 16.94 / 2.22 / 14.49 / 56.07 & 16.17 / 1.91 / 13.92 / 55.86 \\ 
        {Llama3-70B} & 15.72 / 2.02 / 13.63 / 57.64 & 15.44 / 2.12 / 13.38 / 57.71 \\ 
        {Qwen2-72B} & 15.51 / 2.51 / 13.64 / 56.34 & 15.72 / 2.58 / 13.74 / 56.74 \\ 
        \bottomrule

    \end{tabular}
    \caption{Evaluation of LLMs' explanations for correctly identified \unreliable~segments with ensembling two prompts' results. The final scores are calculated by averaging the scores of each explanation generated by the prompts.}
    \label{tab:explanation_score_full_2}
\end{table*}

\section{Prompt Templates}

\label{appendix_prompt_templates}

We provide the detailed prompt templates used for the experiments throughout the paper. This includes prompts for generating LLM reviews (Table~\ref{tab:llm_generate_revivew_prompt}) and identifying \unreliable~segments (Table~\ref{tab:assess_all} and \ref{tab:identify_unreliable}). 

\begin{table*}[h]
    \centering
    \renewcommand{\arraystretch}{1.5}
    \begin{tabular}{>{\columncolor{codebg}}m{0.95\textwidth}}
        Assume you are a meta-reviewer of a natural language processing conference. \\
        Given a paper submission and its corresponding review, your job is to assess the deficiency of each review segment. \\
        The review is segmented, and each segment has an index at the start. You need to assess if each segment of the review is "deficient" or not \\
        The criteria for ``Deficient'' are:\\
        1. Sentences that contain factual errors or misinterpretations of the submission. \\
        2. Sentences lacking constructive feedback. \\
        3. Sentences that express overly subjective, emotional, or offensive judgments, such as ``\textit{I don't like this work because it is written like by a middle school student}.'' \\
        4. Sentences that describe the downsides of the submission without supporting evidence, for example, ``\textit{This work misses some related work}.'' \\
        Your answer should be indexed according to the indices of the segments. For each segment, if it is "reliable," you can simply output "Yes." If it is \unreliable, you should output "No," followed by the reason why it is \unreliable. \\
        In your assessment, consider not only the content of each segment but also the overall context of the review and the paper submission. \\
        Here is the submission title: \\
        \{paper\_title\} \\
        Here is the body text of the submission: \\ 
        \{body\_text\} \\
        Here is the segmented review: \\
        \{review\_text\} \\
        Here is the author rebuttals: \\
        \{author\_rebuttals\_text\} \\
        Your answer should only contain the segment index, your assessment "Yes" or "No," and the explanation if your assessment is "No." Here is an example format: \\
        {[}index{]}. [Yes or No][Your explanation if your answer is No] \\
        Output your answer below:
    \end{tabular}
    \caption{\promptall~ prompt template.}
    \label{tab:assess_all}
\end{table*}

\begin{table*}[h]
    \centering
    \renewcommand{\arraystretch}{1.5}
    \begin{tabular}{>{\columncolor{codebg}}m{0.95\textwidth}}
        Assume you are a meta-reviewer of a natural language processing conference. Given a paper submission and its corresponding review, your job is to assess the deficiency of each review segment. \\
        The review is segmented, and each segment has an index at the start. You need to assess if each segment of the review is "deficient" or not \\
        The criteria for ``Deficient'' are:\\
        1. Sentences that contain factual errors or misinterpretations of the submission. \\
        2. Sentences lacking constructive feedback. \\
        3. Sentences that express overly subjective, emotional, or offensive judgments, such as ``\textit{I don't like this work because it is written like by a middle school student}.'' \\
        4. Sentences that describe the downsides of the submission without supporting evidence, for example, ``\textit{This work misses some related work}.'' \\
        Your answer should include the indices of all \unreliable segments, each followed by the reason why the segment is \unreliable. \\
        In your assessment, consider not only the content of each segment but also the overall context of the review and the paper submission. \\
        Here is the submission title: \\
        \{paper\_title\} \\
        Here is the body text of the submission: \\ 
        \{body\_text\} \\
        Here is the segmented review: \\
        \{review\_text\} \\
        Here is the author rebuttals: \\
        \{author\_rebuttals\_text\} \\
        Your answer should contain only the indices of all \unreliable segments, followed by the reason why each segment is \unreliable. \\
        {[}index{]}. [Your explanation] \\
        Output your answer below:
    \end{tabular}
    \caption{\promptneg~ prompt template.}
    \label{tab:identify_unreliable}
\end{table*}

\begin{table*}[h]
    \centering
    \renewcommand{\arraystretch}{1.5}
    \begin{tabular}{>{\columncolor{codebg}}m{0.95\textwidth}}
        As an esteemed reviewer with expertise in the field of Natural Language Processing (NLP), you are asked to write a review for a scientific paper submitted for publication. Please follow the reviewer guidelines provided below to ensure a comprehensive and fair assessment: \\
        Reviewer Guidelines: 
        \{review\_guidelines\} \\
        In your review, you must cover the following aspects, adhering to the outlined guidelines: \\
        Summary of the Paper: [Provide a concise summary of the paper, highlighting its main objectives, methodology, results, and conclusions.] \\
        Strengths and Weaknesses: [Critically analyze the strengths and weaknesses of the paper. Consider the significance of the research question, the robustness of the methodology, and the relevance of the findings.] \\
        Clarity, Quality, Novelty, and Reproducibility: [Evaluate the paper on its clarity of expression, overall quality of research, novelty of the contributions, and the potential for reproducibility by other researchers.] \\
        Summary of the Review: [Offer a brief summary of your evaluation, encapsulating your overall impression of the paper.] \\
        Correctness: [Assess the correctness of the paper's claims, you are only allowed to choose from the following options: \\
        \{Explanation on different correctness scores\} \\
        Technical Novelty and Significance: [Rate the technical novelty and significance of the paper's contributions, you are only allowed to choose from the following options: \\
        \{Explanation on different Technical Novelty and Significance scores\} \\
        Empirical Novelty and Significance: [Evaluate the empirical contributions, you are only allowed to choose from the following options: \\
        \{Explanation on different Empirical Novelty and Significance scores\} \\
        Flag for Ethics Review: Indicate whether the paper should undergo an ethics review [YES or NO]. \\
        Recommendation: [Provide your recommendation for the paper, you are only allowed to choose from the following options: \\
        \{Explanation on different recommendation scores\} \\
        Confidence: [Rate your confidence level in your assessment, you are only allowed to choose from the following options: \\
       \{Explanation on different confidence scores\} \\
        To assist in crafting your review, here are two examples from reviews of different papers: \\
        \#\# Review Example 1: \\
        \{review\_example\_1\} \\
        \#\# Review Example 2: \\
        \{review\_example\_2\} \\
        Follow the instruction above, write a review for the paper below:
    \end{tabular}
    \caption{Prompt template for generating reviews with LLMs}
    \label{tab:llm_generate_revivew_prompt}
\end{table*}

\end{document}